\RequirePackage{fix-cm}
\documentclass[smallextended]{svjour3}       
\smartqed  
\usepackage{graphicx}
\usepackage{graphicx}
\usepackage{textcomp}
\usepackage{xcolor}

\usepackage{amsfonts, amsmath}
\usepackage{amssymb}
\usepackage{multirow}
\usepackage{multicol}
\usepackage{graphicx}
\usepackage{subfigure}
\usepackage{adjustbox}
\usepackage{array}
\usepackage{booktabs}

\usepackage[ruled,vlined,linesnumbered,boxed]{algorithm2e}
\usepackage{algpseudocode}
\usepackage{bbm}

\newcommand{\argmax}{\operatornamewithlimits{argmax}}

\newtheorem{prob}{Problem}
\begin{document}

\title{Informative Pseudo-Labeling for Graph Neural Networks with Few Labels 
}


\author{Yayong Li \and Jie Yin \and
        Ling Chen 
}


\institute{Yayong Li \at
        University of Technology Sydney \\
        \email{Yayong.Li@student.uts.edu.au}           
           \and
           Jie Yin \at
           The University of Sydney\\
           \email{jie.yin@sydney.edu.au}
           \and
           Ling Chen \at
           University of Technology Sydney\\
           \email{Ling.Chen@uts.edu.au}
}

\date{Received: date / Accepted: date}

\maketitle

\begin{abstract}
Graph Neural Networks (GNNs) have achieved state-of-the-art results for semi-supervised node classification on graphs. 
Nevertheless, the challenge of how to effectively learn GNNs with very few labels is still under-explored. As one of the prevalent semi-supervised methods, pseudo-labeling has been proposed to explicitly address the label scarcity problem. It aims to augment the training set with pseudo-labeled unlabeled nodes with high confidence so as to re-train a supervised model in a self-training cycle. However, the existing pseudo-labeling approaches often suffer from two major drawbacks. First, they tend to conservatively expand the label set by selecting only high-confidence unlabeled nodes without assessing their informativeness. Unfortunately, those high-confidence nodes often convey overlapping information with given labels, leading to minor improvements for model re-training. Second, these methods incorporate pseudo-labels to the same loss function with genuine labels, ignoring their distinct contributions to the classification task. In this paper, we propose a novel informative pseudo-labeling framework, called InfoGNN, to facilitate learning of GNNs with extremely few labels. Our key idea is to pseudo label the most informative nodes that can maximally represent the local neighborhoods via mutual information maximization. To mitigate the potential label noise and class-imbalance problem arising from pseudo labeling, we also carefully devise a generalized cross entropy loss with a class-balanced regularization to incorporate generated pseudo labels into model re-training. Extensive experiments on six real-world graph datasets demonstrate that our proposed approach significantly outperforms state-of-the-art baselines and strong self-supervised methods on graphs. 


\end{abstract}

\section{Introduction}
\noindent Graph neural networks have emerged as state-of-the-art models for undertaking semi-supervised node classification on graphs~\cite{perozzi2014deepwalk}, \cite{kipf2016semi}, \cite{velivckovic2017graph}, \cite{wu2019simplifying}, \cite{hamilton2017inductive}. It aims to leverage a small subset of labeled nodes together with a large number of unlabeled nodes to train an accurate classifier. Most modern GNNs rely on an iterative message passing procedure that aggregates and transforms the features of neighboring nodes to learn node embeddings, which are then used for node classification. However, under extreme cases where very few labels are available (e.g., only a handful of labeled nodes per class), popular GNN architectures (e.g., graph convolutional networks (GCNs) typically with two layers) are ineffective in propagating the limited training labels to learn discriminative node embeddings, resulting in inferior classification performance. Recently, a central theme of latest studies has attempted to improve classification accuracy by designing deeper GNNs or new network architectures~\cite{qu2019gmnn,verma2019graphmix}. However, the challenge of how to effectively learn GNNs with few labels is still under-explored.

Recently, pseudo-labeling, also called self-training, has been proposed as one prevalent semi-supervised method to explicitly tackle the label scarcity problem on graphs. Pseudo-labeling expands the label set by assigning a pseudo label to high-confidence unlabeled nodes, and iteratively re-trains the model with both given labels and pseudo labels. Li et al.~\cite{li2018deeper} first proposed a self-trained GCN that chooses top-$K$ high-confidence unlabeled nodes to enlarge the training set for model re-training. Sun et al.~\cite{Sun2020MultiStageSL} pointed out a shallow GCN's ineffectivenss in propagating label information under few-label settings. A multi-stage approach was then proposed, which applies deep clustering techniques to assign pseudo labels to unlabeled nodes with high prediction confidence. 
Zhou et al.~\cite{zhou2019dynamic} proposed a dynamic self-training framework, which assigns a soft label confidence on the pseudo label loss to control their contribution to gradient update. 

Despite offering promising results, the existing pseudo-labeling approaches on GNNs have not fully explored the power of self-training, due to two major limitations. First, these methods impose strict constraints that only unlabeled nodes with high prediction probabilities are selected for pseudo labeling. However, these selected nodes often convey similar information with given labels, causing \textit{information redundancy} in the expanded label set. 
On the contrary, if unlabeled nodes with lower prediction probabilities are allowed to enlarge the label set, more pseudo label noise would be incurred to significantly degrade the classification accuracy. This creates a dilemma for pseudo-labeling strategies to achieve desirable performance improvements. 
Second, the existing methods all treat pseudo labels and genuine labels equally important. They are incorporated into the same loss function, such as the standard cross entropy loss, for node classification, neglecting their distinct contributions to the classification task. In the presence of unreliable or noisy pseudo labels, model performance might deteriorate during re-training.

Motivated by the above observations, in this paper, we propose a novel informative pseudo-labeling framework called \textbf{InfoGNN} for semi-supervised node classification with few labels. Our aim is to fully harness the power of self-training by incorporating more pseudo labels, but at the same time, alleviate possible negative impact caused by noisy (i.e. incorrect) pseudo labels. To address \textbf{information redundancy}, we define node representativeness via neural estimation of mutual information between a node and its local context subgraph in the embedding space. This method offers two advantages: 1) It provides an informativeness measure to select unlabeled nodes for pseudo labeling, such that the added pseudo labels can bring in more information gain. 2) It implicitly encourages each node to approximate its own local neighborhood and depart away from other neighborhoods. The intuition behind is that an unlabeled node is considered informative when it can maximally reflect its local neighborhood. By integrating 
this informativeness measure with model prediction probabilities, our approach enables to selectively pseudo label nodes with maximum performance gains. To mitigate negative impact of \textbf{noisy pseudo labels}, we also propose a generalized cross entropy  loss on pseudo labels to improve model robustness against noise. This loss allows us to maximize the pseudo-labeling capacity while minimizing the model collapsing risk. In addition, to cope with the potential \textbf{class-imbalance problem} caused by pseudo labeling under extremely few-label settings, we propose a class-balanced regularization that regularizes the number of pseudo labels to keep relative equilibrium in each class. 

Our main contributions can be summarized as follows:
\begin{itemize}
    \item Our study analyzes the ineffectiveness of existing pseudo-labeling strategies and proposes a novel pseudo-labeling framework for semi-supervised node classification with extremely few labels;
    \item Our approach has unique advantages to incorporate an MI-based informativeness measure for pseudo-label candidate selection and to alleviate the negative impact of noisy pseudo labels via a generalized cross entropy loss.
    \item We validate our proposed approach on six real-world graph datasets of various types, showing its superior performance to state-of-the-art baselines.
\end{itemize}

\section{Related works}
\subsection{Graph Learning with Few Labels}
GNNs have emerged as a new class of deep learning models on graphs~\cite{kipf2016semi,velivckovic2017graph}. The principle of GNNs is to learn node embeddings by recursively aggregating and transforming continuous feature vectors from local neighborhoods~\cite{wu2019simplifying,chen2020simple,gao2019graph,chen2018fastgcn}. The generated node embeddings can then be used as input to any differentiable prediction layer, for example, a softmax layer for node classification. Recently, a series of semi-supervised GNNs such as GCNs and their variants have been proposed for node classification. 
The success of these models relies on a sufficient number of labeled nodes for training. How to train GNNs with a very small set of labeled nodes has remained a challenging task. 






\paragraph{\textbf{Pseudo-Labeling on Graphs.}}
To tackle label scarcity, pseudo-labeling has been proposed as one of the prevalent \textit{semi-supervised} methods. It refers to a specific training regime, where the model is bootstrapped with additional labeled data obtained by using a confidence-based thresholding method~\cite{lee2013pseudo,rosenberg2005semi}. 
Recently, pseudo-labeling has shown promising results on semi-supervised node classification. Li et al.~\cite{li2018deeper} proposed a self-trained GCN that enlarges the training set by assigning a pseudo label to top $K$ confidence unlabeled nodes, and then re-trains the model using both given labels and pseudo labels. A co-training method was also proposed that utilizes two models to complement each other. The pseudo labels are given by another random walk model rather than the GNN classifier itself. A similar method was also proposed  in~\cite{zhan2021mutual}. Sun et al.~\cite{Sun2020MultiStageSL} showed that a shallow GCN is ineffective in propagating label information under few-label settings, and proposed a multi-stage self-training framework that relies on a deep clustering model to assign pseudo labels. Zhou et al.~\cite{zhou2019dynamic} proposed a dynamic pseudo-labeling approach called DSGCN that selects unlabeled nodes with prediction probabilities higher than a pre-specified threshold for pseudo labeling, and assigns soft label confidence to them as label weight. 


We argue that all of the existing pseudo-labeling methods on GNNs share two major problems: information redundancy and noisy pseudo labels. Our work is proposed to explicitly overcome these limitations, with a focus on developing a robust pseudo-labeling framework that allows to expand the pseudo label set with more informative nodes, and to mitigate the negative impact of noisy pseudo labels simultaneously.

\paragraph{\textbf{Graph Few-shot Learning.}}
Originally designed for image classification, few-shot learning primarily focuses on the tasks where a classifier is adapted to accommodate new classes unseen during training, given only a few labeled examples for each class~\cite{snell2017prototypical}. Several recent studies~\cite{ding2020graph,huang2020graph,wang2020generalizing} have attempted to generalize few-shot learning to graph domains. For example, Ding et al.~\cite{ding2020graph} proposed a graph prototypical network for node classification, which learns a transferable metric space via meta-learning, such that the model can extract meta-knowledge to achieve good generalization ability on target few-shot classification task. Huang et al.~\cite{huang2020graph} proposed to transfer subgraph-specific information and learn transferable knowledge via meta gradients.

Despite the fact that few-shot learning and our work both tackle the label scarcity problem, their problem settings and learning objectives are fundamentally different: in few-shot learning, the training and test sets typically reside in different class spaces. Hence, few-shot learning aims to learn transferable knowledge to enable rapid generalization to new tasks. On the contrary, our work follows the transductive GNN setting where the training and test sets share the same class space. Our objective is to improve model training in face of very few labels.




\paragraph{\textbf{Graph Self-supervised Learning}}

Our work is related to self-supervised learning on graphs~\cite{velickovic2019deep}, which also investigates how to best leverage the unlabeled data. However, there is a clear distinction in the objectives: the primary aim of self-supervised learning is to learn node/graph representations by designing pretext tasks without label-related supervision, such that the generated representations could facilitate specific classification tasks~\cite{Liu2021GraphSL}. For example,
You et al.~\cite{you2020graph} demonstrated that self-supervised learning can provide regularization for graph-related classification tasks. This work proposed three pretext tasks (i.e., node clustering, graph partitioning, and graph completion) based on graph properties. Other research works attempted to learn better node/graph representations through creating contrastive views, such as local node vs. global graph view in~\cite{velickovic2019deep}, or performing graph augmentation~\cite{zhu2020deep}. 

In contrast, our work resort to explicitly augmenting label-specific supervision for node classification. This is achieved by expanding the existing label set with reliable pseudo labels to best boost model performance in a semi-supervised manner. 

\subsection{Mutual Information Maximization}

The Infomax principal was first proposed to encourage an encoder to learn effective representations that share maximized Mutual Information (MI) with the input~\cite{linsker1988self,belghazi2018mutual,hjelm2018learning}. 
Recently, this MI maximization idea has been applied to improve graph representations. 
Velickovic et al.~\cite{velickovic2019deep} applied MI maximization to learn node embeddings by contrasting local subgraphs and the high-level, global graph representations. 
Qiu et al.~\cite{qiu2020gcc} proposed to learn intrinsic and transferable structural representations by contrasting subgraphs from different graphs via a discriminator. Hassani et al.~\cite{hassani2020contrastive} contrasted node representations from a local view with graph representations from a global view to learn more informative node embeddings. In our context, we leverage the idea of contrastive learning to maximize the MI between each node and its neighboring context. The estimated MI enables to select more representative unlabeled nodes in local neighborhoods for pseudo labeling so as to further advance model performance.

\section{Problem Statement} 
Let $\mathcal{G}=\{\mathcal{V}, \mathcal{E}, X\}$ represents an undirected graph, where $\mathcal{V} = \{v_1, v_2, ...,v_n\}$ denotes a set of $n$ nodes, and $\mathcal{E}$ denotes the set of edges connecting nodes. $X = [\mathbf{x}_1, \mathbf{x}_2, \ldots, \mathbf{x}_n]^T\in \mathbb{R}^{n\times d}$ denotes the node feature matrix, and $\mathbf{x}_i \in \mathcal{R}^d$ is $d$-dimensional feature vector of node $v_i$. The graph structure is represented by the adjacent matrix $A \in \mathbb{R}^{n \times n}$, where $A(i,j) \in \{0, 1\}$. 
We assume that only a small fraction of nodes are labeled in the node set, where $\mathcal{L}=\{(\mathbf{x}_i, \mathbf{y}_i)\}_{i=1}^{|L|}$ denotes the set of labeled nodes, and $\mathcal{U}$ denotes the set of unlabeled nodes. $\mathbf{y}_i=\{y_{i1},y_{i2},\dots,y_{ic}\}$ is the one-hot encoding of node $v_i$'s class label, and $c$ is the number of classes. 

We consider the semi-supervised node classification problem~\cite{kipf2016semi,velivckovic2017graph} under a pseudo-labeling paradigm, which is formally defined as follows:

\begin{prob}
\textit{Given an undirected graph $\mathcal{G}=\{\mathcal{V}, \mathcal{E}, X\}$ together with a small subset of labeled nodes $\mathcal{L}=\{(\mathbf{x}_i, \mathbf{y}_i)\}_{i=1}^{|L|}$, we aim to design a strategy $\mathbbm{1}(\cdot)$ for expanding the label set from unlabeled nodes, a method $\mathbb{Y}(\cdot)$ for generating reliable pseudo labels, and an exclusive loss function $\ell_U(\cdot)$ for pseudo labels, such that $\mathbbm{1}(\cdot)$, $\mathbb{Y}(\cdot)$ and $\ell_U(\cdot)$ can be combined together with the task-specific loss $\ell_L(\cdot)$ to maximize the classification performance of graph neural network $f_{\theta}(\cdot)$. This problem can be formally formulated as Eq.(\ref{defination}).}
\end{prob}

\begin{equation}
\label{defination}
    \min_{\theta} \mathcal{J} = \sum_{\mathbf{x}_i\in \mathcal{L}} \ell_L(\mathbf{y}_{i}, f_\theta(\mathbf{x}_i)) +  \sum_{\mathbf{x}_i\in \mathcal{U}} \ell_U(\mathbf{\mathbb{Y}}(\mathbf{x}_i), f_\theta(\mathbf{x}_i)) \cdot \mathbbm{1}(\mathbf{x}_i).
\end{equation}


\section{Methodology}

\subsection{Framework Overview}
\begin{figure*}
    \centering
    \includegraphics[width=1.0\textwidth]{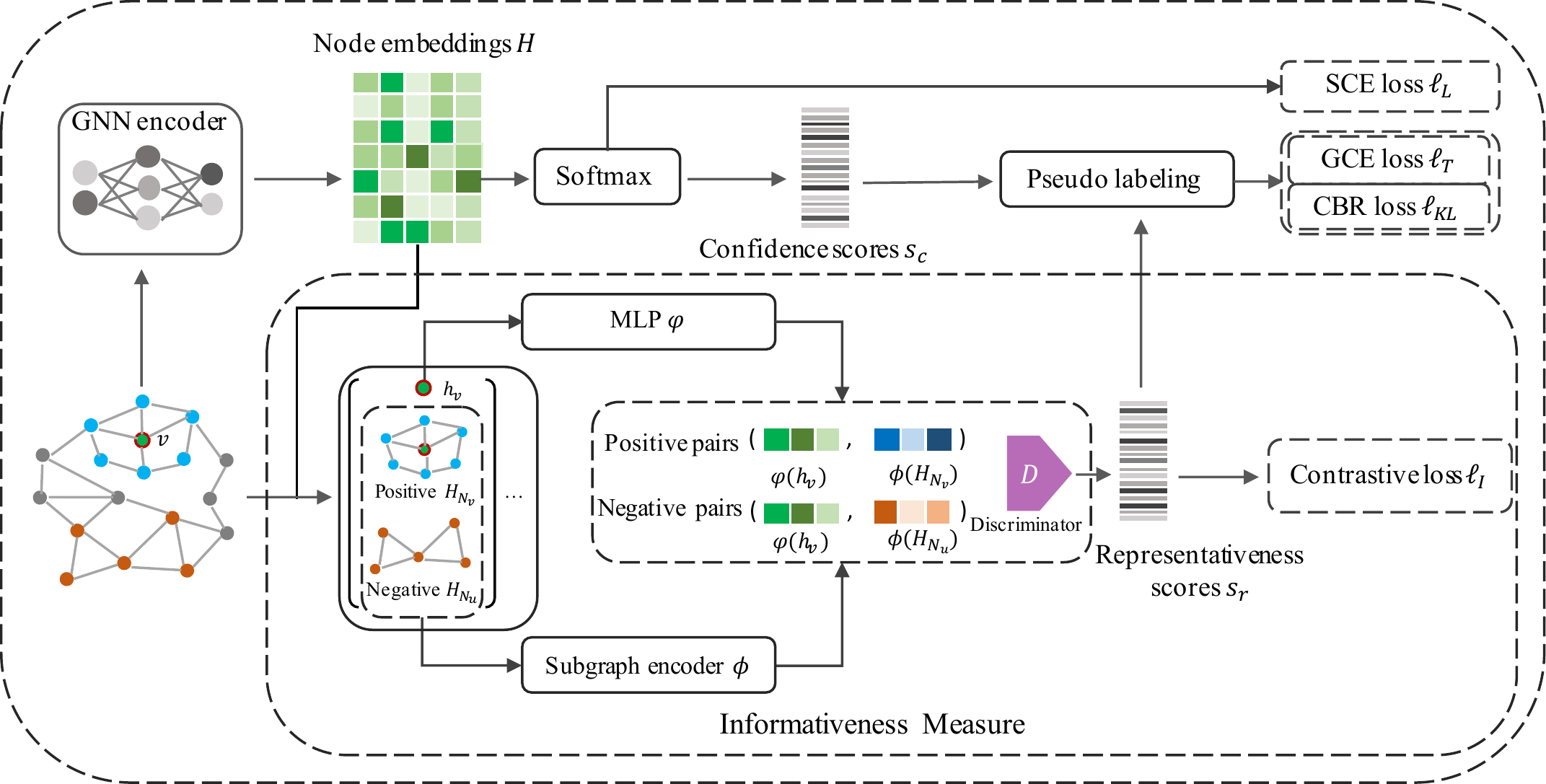}
    \caption{Overview of the proposed informativeness augmented pseudo-labeling framework}
    \label{fig:framework}
\end{figure*}

The primary aim of our work is to develop a robust pseudo-labeling framework for GNN training with few labels. As shown in Fig.~\ref{fig:framework}, our proposed InfoGNN framework comprises of four primary components: 1) a GNN encoder; 2) candidate selection via MI maximization; 3) pseudo labeling; 4) GNN re-training with the generalized cross entropy loss (GCE) and a class-balanced regularization (CBR) on pseudo labels. 

Taking a graph as input, a GNN encoder is first utilized to generate node embeddings and class predictions. Based on the generated node embeddings and graph structure, we derive a measure of node representativenss via MI maximization to assess the informativenss of the nodes, serving for node selection of pseudo labeling. According to the informativeness measure and class prediction probabilities, we assign pseudo labels to selected reliable nodes and use them to augment the existing label set for model re-training. During GNN re-training phase, we propose a GCE loss to improve model robustness against potentially noisy pseudo labels. Furthermore, a KL-Divergence loss is used as a regularizer to mitigate the potential class-imbalanced problem caused by pseudo labeling, which could be exacerbated by the label scarcity problem. Finally, the standard cross entropy loss (SCE) on labeled nodes, the GCE and CBR losses on pseudo labels are integrated to re-train the GNN network.

\subsection{The GNN Encoder}
The GNN encoder is the backbone for our framework. It mainly serves for generating node embeddings and giving class prediction probabilities that  reflect model confidence in terms of predictions. Any GNN that focuses on node classification can be utilized here for embedding learning and classification. A GNN encoder generally learns node embeddings by recursively aggregating and transforming node features from topological neighborhoods. 
In our work, we utilize GCN \cite{kipf2016semi} as our GNN encoder $f_{\theta}(\cdot)$. For $v \in \mathcal{V}$, the node embedding at $k$\textit{-th} layer's prorogation can be obtained by: 
\begin{equation}
\label{eq:GCN}
    h^{k}_{v} =\sigma( \sum_{v'\in \mathcal{N}_v} (\tilde{D}^{-1/2}\tilde{A}\tilde{D}^{-1/2})_{v,v'} W^{k-1}h^{k-1}_{v'}),
\end{equation}
$\sigma(\cdot)$ is the activation function, $\tilde{A} = A + I$ is the adjacency matrix of $\mathcal{G}$ with added self-connections. $\tilde{D}_{ii} = \sum_{j} \tilde{A}_{ij}$, and $W^{k}$ is a layer-specific trainable weight matrix. We use the SCE loss to optimize GCN for node classification: 
\begin{equation}
\label{eq:standardCE}
    \ell_L(\mathbf{y}, f_\theta(\mathbf{x})) = - \sum_{i\in \mathcal{L}} \mathbf{y}_{i} log(f_\theta({\mathbf{x}_i})).
\end{equation}
Finally, according to the class prediction probabilities, we can obtain the \textit{confidence score} for each node $v$:
\begin{equation} \label{eq:poolup}
         s_c(v) = \max_j f_{\theta}(\mathbf{x}_v)_j,
\end{equation}
The confidence score $s_c(v)$ is utilized for node selection in combination with the representativeness score, which is detailed below. 

\subsection{Candidate Selection for Pseudo Labelling}
Most existing pseudo-labeling methods typically select unlabeled nodes accounting for only model confidence or uncertainty~\cite{zhou2019dynamic,li2018deeper,mukherjee2020uncertainty}. These methods are in favor of selecting the nodes with high prediction probabilities, expecting to bring in less noisy pseudo labels for model re-training. However, these high-confidence nodes tend to carry redundant information with given labels, resulting in limited capacity to improve model performance. Therefore, besides model confidence, we propose to take node informativeness into account for node selection so as to maximally boost model performance. To this end, the key problem lies in how to measure node informativeness. 

\paragraph{\textbf{Informativeness Measure by MI Maximization}}
We define the node informativeness as the representativeness of a node in relation to its contextual neighborhood. The intuition behind is that a node is informative when it could maximally represent its surrounding neighborhood while minimally reflect other arbitrary neighborhoods. 
Hence, the representativeness of a node can be measured by the mutual information between the node and its neighborhood with positive correlation. On account of this, we employ \textit{MI maximization} techniques~\cite{belghazi2018mutual} to estimate the MI by measuring how much one node can represent its surrounding neighborhood and discriminate an arbitrary neighborhood. This thus provides a score to quantify the representativeness for each node. We achieve this by formulating it to be a \textit{subgraph-based contrastive learning} task, which contrasts each node with its positive and negative context subgraphs. 


Given a graph $\mathcal{G}=\{\mathcal{V}, \mathcal{E}, X\}$ with learned node embeddings $H$, for each node $v \in \mathcal{V}$, we define its local $r$-hop subgraph $\mathcal{N}_v$ as the \textit{positive} sample, and an arbitrary $r$-hop subgraph $\mathcal{N}_u$ from node $u$ as the \textit{negative} sample. The mutual information $I^\mathcal{G}(v)$ between node $v$ and its neighborhood can then be measured by a GAN-like divergence~\cite{nowozin2016f}:

\begin{equation}
\begin{split}
    I^\mathcal{G} \geqslant \hat{I}^\mathcal{G} &= \max_{\omega} \frac{1}{|\mathcal{V}|}\sum_{v\in\mathcal{V}}[MI_{\omega}(h_v, H_{\mathcal{N}_v}) + MI_{\omega}(h_v, H_{\mathcal{N}_u})]
\end{split}
\label{eq:mi}
\end{equation}
where $h_v$ is node $v$'s embedding generated from the GNN encoder, $H_{\mathcal{N}_v}$ and $H_{\mathcal{N}_u}$ are the embedding sets of subgraphs centered at node $v$ and $u$.  $MI_{\omega}$ is the trainable neural network parameterized by $\omega$. $MI_{\omega}(h_v, H_{\mathcal{N}_v})$ indicates the affinity between positive pairs, while $MI_{\omega}(h_v, H_{\mathcal{N}_u})$ indicates the discrepancy between negative pairs. Our objective is to estimate $I^\mathcal{G}$ by maximizing $MI_{\omega}(h_v, H_{\mathcal{N}_v})$ while minimizing $MI_{\omega}(h_v, H_{\mathcal{N}_u})$, which is in essence a contrastive objective. 

This contrastive objective is further achieved by employing a discriminator as illustrated in Figure~\ref{fig:framework}. At each iteration, after obtaining the learned node embeddings $H$, both positive and negative subgraphs for each node are firstly sampled and paired. Then those nodes and their corresponding paired subgraphs are passed on to a discriminator $\mathcal{D}(\cdot)$ after being separately processed by an MLP encoder $\varphi(\cdot)$ and a subgraph encoder $\phi(\cdot)$. This discriminator finally produces a representativeness score for each node by distinguishing a node's embedding from its subgraph embedding.

Formally, we specify $MI_{\omega}(h_v, H_{\mathcal{N}_v}) = \mathcal{D}(\varphi(h_v), \phi(H_{\mathcal{N}_v}))$. Here, $\varphi(\cdot)$ is an MLP encoder for node embedding transformation. 
$\phi(\cdot)$ is a \textit{subgraph encoder} that aggregates embeddings of all nodes in the subgraph to generate an embedding of the subgraph, which is implemented using a \textit{one-layer GCN} on a $r$-hop subgraph:
\begin{equation}
\begin{split}
    &\phi(H_{\mathcal{N}_v}) = \sum_{v'\in \mathcal{N}_v} (D_r^{-1/2}A_{r}D_r^{-1/2})_{v,v'} h_{v'}W,\\
    & A_r = Bin(A*A_{r-1} + A_{r-1})
\end{split}
\end{equation}
where $A$ is the original graph adjacent matrix, and $Bin(\cdot)$ is a binary function that guarantees $A_r(i,j) \in \{0, 1\}$. $D_r$ is the corresponding degree matrix of $A_r$. What we need to notice is that we use a $r$-hop adjacent matrix $A_r$, instead of the original adjacent matrix $A$ for feature aggregation, and the aggregated embedding of the centre node in the subgraph will be the subgraph embedding. With regard to the discriminator $\mathcal{D}(\cdot)$, we implement it using a bilinear layer:
\begin{equation}
    \mathcal{D}(\varphi(h_v), \phi(H_{\mathcal{N}_v})) = \sigma(\varphi(h_v)B\phi(H_{\mathcal{N}_v})^T)
\end{equation}
where $B$ is the learnable parameter. 
To enable the discriminator $\mathcal{D}(\varphi(h_v), \phi(H_{\mathcal{N}_v}))$ to measure the affinity between node $v$ and its corresponding local subgraph $\mathcal{N}_v$, we minimize the binary cross entropy loss between positive and negative pairs, which is formulated as the \textit{contrastive loss}:
\begin{equation}
\begin{split}
    \ell_{I} = & - \frac{1}{|\mathcal{V}|}\sum_{v\in\mathcal{V}}[log \mathcal{D}(\varphi(h_v), \phi(H_{\mathcal{N}_v}))
    +log(1-\mathcal{D}(\varphi(h_v), \phi(H_{\mathcal{N}_u})))],
\end{split}
\end{equation}


By minimizing $\ell_I$, the discriminator could maximally distinguish a node from any arbitrary subgraphs that it does not belong to in the embedding space. This process is equivalent to maximizing their MI in the sense of Eq.(\ref{eq:mi}). 

\paragraph{\textbf{Pseudo Labeling.}} The discriminator $\mathcal{D}(\cdot)$ measures the affinity between each node and its local subgraph. We utilize this affinity to define the \textit{informativeness score} for each node: 
\begin{equation}
\label{eq:score}
    s_r(v) = \mathcal{D}(\varphi(h_v), \phi(H_{\mathcal{N}_v})),
\end{equation}
where $s_r(v)$ indicates to what extent a node could reflect its neighborhood, and a higher score means that the node is more informativeness. Therefore, by considering both the informativeness score and model prediction confidence, we derive the selection criterion to construct the pseudo-label set $U_p$:
\begin{equation} 
\label{eq:up}
         U_p = \{v \in \mathcal{U} | (s_r(v) + s_c(v)) / 2 > k, s.t.  s_c(v) > k\}.
\end{equation}
where $s_c(v)$ is the confidence score as in Eq.(\ref{eq:poolup}), and $k$ is a hyperparameter whose value can be empirically determined (See Fig.~\ref{fig:kw1} in Section~\ref{subsection: sensitivity}). We then produce the pseudo labels for $U_p$ utilizing the GNN encoder $f_{\theta}(\cdot)$:
\begin{equation}
    \hat{\mathbf{y}}_v = \argmax_j f_{\theta}(\mathbf{x}_v)_j; v \in U_p
\end{equation}
Where the pseudo label $\hat{\mathbf{y}}_v$ is actually the predicted label by the GNN encoder. 





\subsection{Mitigating Noisy Pseudo Labels}

During the re-training phase, existing pseudo-labeling methods regard given labels and pseudo labels equally important, so an identical loss function, e.g., the SCE loss, is applied. However, with more nodes added in, it is inevitable to introduce unreliable or noisy (i.e., incorrect) pseudo labels. If the same SCE loss is still applied on unreliable pseudo labels, it would degrade model performance. This is because, the SCE loss implicitly weighs more on the difficult nodes whose predictions deviate away from the supervised labels during gradient update~\cite{zhang2018generalized,van2015learning}. This is beneficial for training with clean labels and ensures faster convergence. However, when there exist noisy pseudo labels in the label set, more emphasis would be put on noisy pseudo labels as they are harder to fit than correct ones. This would ultimately cause the model to overfit incorrect labels, thereby degrading model performance. 

To address this issue, we propose to apply the negative Box-Cox transformation~\cite{box1964analysis} to the loss function $\ell_U(\cdot)$ on pseudo label set $U_p$, inspired by~\cite{zhang2018generalized}. The transformed loss function is given as follows:
\begin{equation}
\label{loss U}
\begin{split}
    \ell_U(\hat{\mathbf{y}}_i, f_{\theta}(\mathbf{x}_i)) &= \frac{1-f_{\theta}(\mathbf{x}_i)^q_j}{q}, \mathbf{x}_i \in U_p,\\
    \hat{\mathbf{y}}_i &= \argmax_j f_{\theta}(\mathbf{x}_i)_j
\end{split}
\end{equation}
where $q \in (0, 1]$, $\hat{\mathbf{y}}_i$ is the pseudo label. To further elaborate how this loss impacts parameter update, we have its gradient as follows:
\begin{equation}
\begin{split}
    \frac{\partial \ell_U(\hat{\mathbf{y}}_{i}, f_{\theta}(\mathbf{x}_i))}{\partial \theta} =  f_{\theta}(\mathbf{x}_i)_j^q (-\frac{1}{f_{\theta}(\mathbf{x}_i)_j}\nabla_{\theta}f_{\theta}(\mathbf{x}_i)),
\end{split}
\end{equation}
where $f_{\theta}(\mathbf{x}_i)_j \in (0, 1]$ for $\forall i$. Compared with the SCE loss, it actually weighs each gradient by an additional $f_{\theta}(\mathbf{x}_i)_j^q$, which reduces the gradient descending on those unreliable pseudo labels with lower prediction probabilities. Actually, $\ell_U(\hat{\mathbf{y}}_i, f_{\theta}(\mathbf{x}_i))$ can be regarded as the \textit{generalization of the SCE loss and the unhinged loss}. It is equivalent to SCE when $q$ approaches zero, and becomes the unhinged loss when $q$ is equal to $one$. Thus, this loss allows the network to collect more additional information from a larger amount of pseudo labels while alleviating their potential negative effect. 



In practice, we apply a truncated version of $\ell_U(\cdot)$ to filter out potential impact from unlabeled nodes with low prediction probabilities, given by:
\begin{equation}
    \ell_T(\hat{\mathbf{y}}_i, f_{\theta}(\mathbf{x}_i)) = \left\{\begin{matrix}
 \ell_U(k),& f_{\theta}(\mathbf{x}_i)_j \leq k\\ 
 \ell_U(\hat{\mathbf{y}}_i, f_{\theta}(\mathbf{x}_i)),& f_{\theta}(\mathbf{x}_i)_j > k
\end{matrix}\right.
\end{equation}
where $k \in (0, 1)$, and $\ell_U(k) = (1-k^q)/q$. Formally, the truncated loss version is derived as:
\begin{equation}
\label{eq:loss_T1}
\begin{split}
    \ell_T(\hat{\mathbf{y}}, f_{\theta}(\mathbf{x}))
     = \sum_{i\in \mathcal{U}} \lambda _i\ell_U(\hat{\mathbf{y}}_i, f_{\theta}(\mathbf{x}_i))+(1-\lambda _i)\ell_U(k),
\end{split}
\end{equation}
where $\lambda _i = 1$ if $i \in U_p$, otherwise $\lambda _i = 0$. Intuitively, when the prediction probability of one node is lower than $k$, the corresponding truncated loss would be a constant. As the gradient of a constant loss is zero, this node would have no contribution to gradient update, thus eliminating negative effect of pseudo labels with low confidence. 




\subsection{Class-balanced Regularization}

Under extreme cases where only very few labels are available for training, severe class-imbalance problem would occur during pseudo labeling. That means, one or two particular classes might dominate the whole pseudo label set, thus conversely impacting model re-training. To mitigate this, we propose to apply a class-balanced regularizer that prevents the number of different classes deviating a lot from each other. For this purpose, we apply a KL-divergence between the pseudo label distribution and a default label distribution:
\begin{equation}
\label{eq:loss_k}
    \ell_{KL} = \sum_{j=1}^c p_j log\frac{p_j}{\overline{f({X})}_j}, 
\end{equation}
where $p_j$ is the default probability of class $j$. Since it would be desirable to have roughly equal numbers of pseudo labels from each class, we set the default label distribution as $p_j = 1/c$ in our situation. $\overline{f({X})}_j$ is the mean value of prediction probability distribution over pseudo labels, which is calculated as follows:
\begin{equation}
\label{eq:prior}
    \overline{f({X})}_j = \frac{1}{|U_p|}\sum_{\mathbf{x}_i \in U_p}f(\mathbf{x}_i)_j
\end{equation}

\subsection{Model Training and Computational Complexity}
Our proposed InfoGNN framework is given by Algorithm~\ref{alg:pseudo code}, which consists of one pre-training phase and one formal training phase. The pre-training phase (Step 2-4) is used to train a parameterized GNN with given labels. Accordingly, network parameters are updated by:
\begin{equation}
\label{eq:pre}
    \ell_{pre}=\ell_L + \alpha \ell_I
\end{equation} 
At the beginning of the formal training phase, the pre-trained GNN is applied to generate prediction probabilities and informativeness score for each node, which are then used to produce pseudo labels (Step 6-8). Finally, both given labels and pseudo labels are used to re-train the GNN by minimizing the following loss function (Step 9):
\begin{equation}
\label{eq:overall}
    \ell = \ell_L + \ell_T + \alpha\ell_I + \beta\ell_{KL}
\end{equation}

\vspace{-0.5cm}
\begin{algorithm}[th]
  \caption{Training InfoGNN with few labels}
  \label{alg:pseudo code}
  \KwIn{Graph $G=\{\mathcal{V}, \mathcal{E}, \mathbf{X}\}$,  $\alpha$, $\beta$, $r, q$ and $k$} 
  \KwOut{label predictions}
  Initialize network parameters\; 
  
  \For{$t=0; t < epoches; t=t+1$}
  {
  \If{$t < $ start\_epoch}
  {
  pre-train the network according to  Eq.(\ref{eq:pre}); 
  }
  \Else
  {
  Generate node prediction probabilities $f_{\theta}(\mathbf{x}_i)$\;
  Generate informativeness scores based on Eq.(\ref{eq:score});
  
  Construct pseudo label set based on Eq.(\ref{eq:up});
  
  Update network parameters based on Eq.(\ref{eq:overall});
  
  }
  }
 \Return{Label predictions}
\end{algorithm}
\vspace{-0.5em}
In terms of computational complexity, by comparison with GNN models based on the SCE loss, InfoGNN incurs slightly extra computational overhead in its attempt to mitigate label noise. The is mainly due to the calculation of the contrastive loss $\ell_I$ with subgraph encoder. Since we utilize a one-layer GCN as subgraph encoder on a $r$-hop subgraph, its computational complexity is linear with the number of edges $\mathcal{O}(|E_r|)$, where $E_r$ is the number of edges in the $r$-hop subgraph, i.e. $|E_r| = sum(A_r)$. This is reasonably acceptable.

\section{Experiments}
To validate the effectiveness of the proposed pseudo-labeling framework, we carry out extensive experiments on six real-world graph datasets to compare against state-of-the-art baselines. We also conduct ablation study and sensitivity analysis to better understand key ingredients of our approach. 

\subsection{Datasets}
Our experiments use six benchmark graph datasets in three different domains: 
\begin{itemize}
\item \textbf{Citation networks:} Cora, Citeseer\cite{kipf2016semi} and Dblp\footnote{https://github.com/abojchevski/graph2gauss}\cite{bojchevski2017deep}. On the three networks, each node represents a paper with a certain label and each edge represents the citation links between two papers. Node features are bag-of-words vectors of papers. 

\item \textbf{Webpage networks:} Wikics\footnote{https://github.com/pmernyei/wiki-cs-dataset/raw/master/dataset}\cite{mernyei2020wiki} is computer science related Webpage networks in Wikipedia. Nodes represent articles about computer science, and edges represent hyperlinks between articles. The features of nodes are mean vectors of GloVe word embeddings of articles.

\item \textbf{Coauther networks}: Coauther-CS and Coauther-Phy\footnote{https://github.com/shchur/gnn-benchmark}\cite{shchur2018pitfalls}. They are coauthor networks in the domain of computer science and Physics. Nodes are authors, and edges mean whether two authors coauthored a paper. Node features are paper keywords from the author’s papers. 
\end{itemize}
Detailed dataset statistics are listed in Table~\ref{table:dataset} below. 

\vspace{-0.5cm}
\begin{table}[h]
    \centering
    \caption{Details of Five Benchmark Datasets}
    \scalebox{1.0}{
    \begin{tabular}{ccccc}
         \toprule
         Dataset & Nodes & Edges & Classes & Features\\
         \midrule
         Citeseer & 3327 & 4732 & 6 & 3703  \\
         Cora & 2708 & 5429 & 7 & 1433 \\
         Dblp & 17716 & 105734 & 4 & 1639 \\
         Wikics & 11701 & 216123 & 10 & 300 \\
         Coauthor\_CS & 18333 & 81894 & 15 & 6805  \\
         Coauthor\_Phy & 34493 & 247962 & 5 & 8415 \\
    \bottomrule
    \end{tabular}}
    \label{table:dataset}
\end{table}
\vspace{-1.2cm}

\subsection{Baselines}
For comparison, we use 12 representative methods as our baselines. Since all methods are based on the original GCN, GCN~\cite{kipf2016semi} is selected as the benchmark. A total of 11 recently proposed methods on graphs are used as strong competitors, which can be categorized into two groups:
\begin{itemize}

\item \textbf{Pseudo-labeling methods}:
M3S~\cite{Sun2020MultiStageSL}, Self-training~\cite{li2018deeper}, Co-training~\cite{li2018deeper}, Union~\cite{li2018deeper}, Intersection~\cite{li2018deeper}, and DSGCN~\cite{zhou2019dynamic};

\item \textbf{Self-supervised methods}: Super-GCN~\cite{Kim2021HowTF}, GMI~\cite{peng2020graph},  SSGCN-clu~\cite{You2020WhenDS},  SSGCN-comp~\cite{You2020WhenDS}, SSGCN-par~\cite{You2020WhenDS}.
\end{itemize}
We run all experiments for 10 times with different random seeds, and report their mean Micro-F1 scores for comparison. 


\subsection{Experimental setup}

\paragraph{Model Specification.} For fair comparison, all baselines are adapted to use a two-layer GCN with 16 units of hidden layer. The hyper-parameters are the same with the GCN in~\cite{kipf2016semi}, with L2 regularization of $5*10^{-4}$, learning rate of 0.01, dropout rate of 0.5. As for subgraph encoder $\phi(\cdot)$, we utilize a one-layer GCN with $c$ outputs, where $c$ is the number of classes. Both positive and negative subgraphs share the same subgraph encoder. $\varphi(\cdot)$ is also a one-layer MLP with 16-dimension output. The discriminator $\mathcal{D}(\cdot)$ is a 1-layer bilinear network with one-dimension output. For dataset split, we randomly choose $\{1, 3, 5, 10, 15, 20\}$ nodes per class for training as different settings. 
Then we randomly pick 30 nodes per class as validation set, and the remaining nodes are used for testing. 
\vspace{-0.5em}
\paragraph{Hyperparameter Specification.}
We specify hyperparameters conforming to the following rules:
\vspace{-0.5cm}
\begin{table}[h]
    \centering
    \caption{Details of hyperparameters}
    \scalebox{1.0}{
    \begin{tabular}{ccccc}
         \toprule
         setting & $\alpha$ & $\beta$ & $q$ & $k$\\
         \midrule
         $lr=\{1,3,5\}$ & 1.0 & 1.0 & 1.0 & 0.55  \\
         $lr=\{10,15,20\}$ & 0.2 & 0.2 & 0.1 & 0.55 \\
         \bottomrule
    \end{tabular}}
    \label{table:hyperparam}
\end{table}
\vspace{-1em}
Generally, a larger $\alpha$ and $\beta$ would be beneficial to model training when given labels are very scarce, while it is more likely to achieve better performance with a smaller $\alpha$ and $\beta$ as the number of given labels increases. We empirically find that our model has relatively lower sensitivity to $q$ with the regularization of loss $\ell_I$, so we can fix its value for most of the settings. Specifically, we utilize $q=1.0$ when 1 label per class are given, and $q=0.1$ for all of the other situations. As for $k$, we fix it to be 0.55 for all the settings.
The best $r$ for subgraph embedding in loss $\ell_I$ depends on the edge density of the input graph. We apply $r=3$ for edge-sparse graphs \{Cora, Citeseer, Dblp, Coauther\_cs\}, $r=2$ for Wikics and $r=1$ for Coauther\_phy.
\vspace{-0.5em}
\paragraph{Implementation Details.}
When training InfoGNN, we first pre-train the network to generate reliable predictions using Eq.(\ref{eq:pre}) for 200 epoches, and then proceed with formal training using the full loss function Eq.(\ref{eq:overall}) for another 200 epoches. During formal training, in order to get a steady model, we allow the model to update pseudo-label set every 5 epoches using Eq.(\ref{eq:up}). When updating the pseudo-label set, we use the mean score of unlabeled nodes in its last 10 training epoches, rather than the current prediction and informativeness score. Our framework is implemented using Pytorch. All experiments are run on a machine powered by Intel(R) Xeon(R) Gold 6126 @ 2.60GHz CPU and 2 Nvidia Tesla V100 32GB Memory Cards with Cuda version 10.2.


\begin{table*}[h]
    \centering
    \tabcolsep 3pt
    \caption{The Micro-F1 performance comparison with various given labels. `-' indicates the method does not work on the dataset, and OOM indicates Out-Of-Memory on a 32GB GPU}
    \scalebox{0.85}{
        \begin{tabular}{c|cccccc|cccccc}
             \toprule
             \multicolumn{1}{ c| }{Method} & \multicolumn{6}{ c| }{Cora} & \multicolumn{6}{ c }{Citeseer} \\
             \cline{2-13} &1& 3 &5 & 10 &15 &20 &1 & 3 &5 & 10 &15  &20  \\
             \midrule
             GCN  &0.416 &0.615 &0.683 &0.742 &0.784& 0.797 
             & 0.379 & 0.505 & 0.568 &0.602& 0.660 & 0.682\\
             
             Super-GCN  &0.522 &0.673 &0.720 &0.760 &0.788 & \underline{0.799} 
             &\underline{0.499} & 0.610 & \underline{0.665} & \underline{0.700} & \underline{0.706} & 0.712\\
             GMI  &0.502 &0.672 &0.715 &0.757 &0.783 & 0.797 
             & 0.497 & 0.568 & 0.621 &0.632 & 0.670 & 0.683 \\
             SSGCN-clu  &0.407 &0.684 &0.739 &0.776 &\underline{0.797} & \underline{0.810} 
             & 0.267 & 0.388 & 0.507 &0.616 & 0.634 & 0.647\\
             SSGCN-comp  &0.451 &0.609 &0.676 &0.741 &0.772 & 0.794 
             & 0.433 & 0.547 & 0.638 &0.682 & 0.692 & 0.709\\
             SSGCN-par  &0.444 &0.649 &0.692 &0.734 &0.757 & 0.770 
             & 0.457 & 0.578 & 0.643 &0.693 & 0.705 & \underline{0.716}\\
             
             Cotraining  &0.533 &0.661 &0.689 &0.741 &0.764& 0.774 
             & 0.383 & 0.469 & 0.563 &0.601& 0.640 & 0.649\\
             Selftraining  &0.399 &0.608 &0.693 &0.761 &0.789& 0.793 
             & 0.324 & 0.463 & 0.526 &0.647& 0.683 & 0.685\\
             Union  &0.505 &0.663 &0.713 &0.764 &0.792& 0.797 
             & 0.366 & 0.491 & 0.560 &0.631& 0.663 & 0.667\\
             Intersection  &0.408 &0.596 &0.674 &0.736 &0.770& 0.775 
             & 0.337 & 0.497 & 0.582 &0.671& {0.694} & {0.699}\\
             M3S  &0.439 &0.651 &0.688 &0.754 &0.763 & 0.789 
             & 0.307 & 0.515 & 0.635 &{0.674} & 0.683 & 0.695\\
             DSGCN  &\underline{0.596} &\underline{0.712} &\underline{0.745} &\underline{0.777} &0.792 & 0.795 
             & {0.463} & \underline{0.613} & {0.652} &{0.674} & 0.681 & 0.684\\
             InfoGNN &\textbf{0.602} &\textbf{0.737} &\textbf{0.775} &\textbf{0.792} &\textbf{0.814} & \textbf{0.829} 
             & \textbf{0.541} & \textbf{0.654} & \textbf{0.717} &\textbf{0.723} & \textbf{0.725} & \textbf{0.734} \\
             \bottomrule
               \multicolumn{1}{ c| }{Method} & \multicolumn{6}{ c| }{Dblp} & \multicolumn{6}{ c }{Wikics} \\
             \cline{2-13} &1& 3 &5 & 10 &15 &20 &1 & 3 &5 & 10 &15  &20  \\
             \midrule
             GCN  &0.469 &0.583 &0.628 &0.652 &0.688 & 0.718 
             & 0.384 & 0.548 & 0.639 &0.682& 0.713 & 0.721\\
             
             Super-GCN  &0.472 &0.583 &0.685 &0.708 &0.729 & 0.738 
             &0.399 & 0.552 & 0.599 &0.683 & 0.712 & 0.721\\
             GMI  &0.544 &0.597 &0.656 &0.728 &0.739 & 0.754 
             & 0.325 & 0.484 & 0.546 &0.654 & 0.683 & 0.700 \\
             SSGCN-clu  &0.369 &0.528 &0.649 &0.692 &0.721 & 0.744 
             & 0.335 & 0.579 & 0.627 &0.694 & 0.714 & 0.725\\
             SSGCN-comp  &0.458 &0.525 &0.598 &0.634 &0.674 & 0.707 
             & - & - & - &- & - & - \\
             SSGCN-par  &0.418 &0.545 &0.639 &0.683 &0.708 & 0.733 
             & 0.332 & 0.593 & 0.659 &0.706 & 0.732 & 0.740\\
             
             Cotraining  &0.545 &0.646 &0.634 &0.674 &0.703 & 0.701 
             & 0.367 & 0.584 & 0.645 &0.692& 0.724 & 0.737\\
             Selftraining  &0.437 &0.580 &0.634 &0.707 &0.738& 0.759 
             & 0.350 & 0.602 & \textbf{0.655} &0.701& 0.725 & 0.738\\
             Union  &0.485 &0.618 &0.652 &0.712 &0.737 & 0.746 
             & 0.351 & 0.584 & 0.646 &0.694& 0.723 & \underline{0.740}\\
             Intersection  &0.458 &0.581 &0.566 &0.665 &0.715 & 0.734 
             & 0.359 & 0.599 & 0.654 &\underline{0.706}& \underline{0.726} & \underline{0.740}\\
             M3S  &0.547 &0.635 &0.672 &0.733 &\underline{0.749} & 0.752 
             &0.401 & 0.593 & 0.621 & 0.685 &0.711 & 0.734 \\
             DSGCN  &\underline{0.587} &\textbf{0.671} &\underline{0.720} &\underline{0.738} &0.744 & \underline{0.764} 
             &\underline{0.414} & \underline{0.607} & 0.635 & 0.705 &0.716 & 0.728 \\
             InfoGNN &\textbf{0.597} &\underline{0.669} &\textbf{0.748} &\textbf{0.765} &\textbf{0.772} & \textbf{0.789} 
             & \textbf{0.462} & \textbf{0.611} & \underline{0.649} &\textbf{0.723} & \textbf{0.740} & \textbf{0.742} \\
             \bottomrule
               \multicolumn{1}{ c| }{Method} & \multicolumn{6}{ c| }{Coauther\_cs} & \multicolumn{6}{ c }{Coauther\_phy} \\
             \cline{2-13}&1& 3 &5 & 10 &15 &20 &1 & 3 &5 & 10 &15  &20  \\
             \midrule
             GCN  &0.642 &0.800 &0.847 &0.893 &0.901 & 0.909 
             & 0.699 & {0.851} & {0.868} &{0.901} & {0.912} & {0.918}\\
             
             Super-GCN  &0.668 &0.841 &0.869 &0.895 &0.897 & 0.897 
             &0.688 & 0.848 & 0.891 &0.908 & 0.923 & 0.923\\
             GMI  &OOM &OOM &OOM &OOM &OOM & OOM 
             & OOM & OOM & OOM &OOM & OOM & OOM\\
             SSGCN-clu  &\textbf{0.770} &\textbf{0.886} &\underline{0.890} &\underline{0.905} &0.908 & 0.911 
             & \textbf{0.889} & \underline{0.923} & \underline{0.930} &\underline{0.935} & \underline{0.936} & \underline{0.936} \\
             SSGCN-comp  &0.711 &0.858 &0.888 &0.904 &0.907 & 0.909 
             & 0.798 & 0.892 & 0.904 &0.927 & 0.921 & .928\\
             SSGCN-par  &0.737 &0.860 &0.881 &0.898 &0.901 & 0.903 
             & 0.824 & 0.915 & 0.919 &0.925 & 0.931 & 0.931\\
             
             Cotraining  &0.643 &0.745 &0.810 &0.849 &0.864 & 0.885
             & 0.758 & 0.842 & 0.850 &0.898 & 0.891 & 0.917\\
             Selftraining  &0.592 &0.770 &0.828 &0.873 &0.892 & 0.895 
             & 0.744 & 0.865 & 0.890 &{0.908} & 0.914 & 0.921\\
             Union  &0.621 &0.772 &0.812 &0.856 &0.864 & 0.885 
             & 0.750 & 0.855 & 0.870 &{0.908} & 0.902 & 0.910\\
             Intersection  &0.650 &0.775 &0.851 &0.887 &0.893 & 0.898 
             & 0.612 & 0.763 & 0.854 &0.901 & 0.904 & 0.926\\
             M3S  &0.648 &0.818 &{0.879} &{0.897} &\underline{0.909} & \underline{0.912} 
             &{0.828} & {0.868} & {0.895} & 0.914 &{0.922} & {0.930} \\
             DSGCN  &\underline{0.743} &{0.829} &0.863 &0.879 &0.883 & 0.892 
             & 0.781 & 0.812 & 0.862 &0.896& 0.908 & 0.916\\
             InfoGNN &{0.682} &\underline{0.866} &\textbf{0.892} &\textbf{0.906} &\textbf{0.913} & \textbf{0.918} 
             & \underline{0.842} & \textbf{0.924} & \textbf{0.934} &\textbf{0.938} & \textbf{0.942} & \textbf{0.942} \\
             \bottomrule
        \end{tabular}}
    \label{table:overall1}
\end{table*}

\vspace{-0.5em}
\subsection{Comparison with State-of-the-art Baselines}
\label{experiment:comparison}
Table \ref{table:overall1} reports the mean Micro-F1 scores of our method and all baselines with respect to various label rates. The best performer is highlighted by \textbf{bold}, and the second best performer is highlighted by \underline{underline} on each setting. 
On the whole, our proposed InfoGNN algorithm outperforms other baseline methods by a large margin over almost all the settings. Compared with GCN, we averagely achieve $12.3\%, 9.3\%, 8.0\%, 6.3\%, 4.1\%, 3.5\%$ of performance improvement on the six datasets when $1, 3, 5, 10, 15, 20$ nodes per class are labeled, respectively. With the help of our pseudo-labelling method, InfoGNN can achieve better results with less nodes. Particularly, by leveraging less 10 nodes per class, InfoGNN succeeds to achieve the similar Micro-F1 scores that GCN achieves using 20 nodes per class over the six datasets. With regard to the self-supervised methods, they have unstable performances over different datasets. For example, although SSGCN-clu obtains the advantageous results on Coauthor-cs/phy datasets, it has relatively worse results on other four datasets. We can also find the SSGCN-Comp even does not work on Wikics dataset. This is because their pretext tasks are probably not always able to help generate representations that generalize well on graphs with different properties. From the table we can also see that, although most baseline methods gradually lose their edges as more labels are available for training, our proposed method still achieves relatively better classification accuracy. Taking 20 labels per class as an example, when all baselines hardly improve classification accuracy over GCN, our proposed method still achieves further improvements. This proves that our algorithm is able to effectively relieve the information redundancy problem when label information is relatively sufficient.

\vspace{-0.5cm}
\begin{table*}[h]
    \centering
    \tabcolsep 3pt
    \caption{The Micro-F1 performance comparisons with various ablation studies }
    \scalebox{0.85}{
        \begin{tabular}{c|cc|cc|cc|cc|cc|cc}
             \toprule
             Method & \multicolumn{2}{c|}{Cora} & \multicolumn{2}{c|}{Citeseer}& \multicolumn{2}{c|}{Dblp}&
             \multicolumn{2}{c|}{Wikics}&\multicolumn{2}{c|}{Coauther\_cs}& \multicolumn{2}{c}{Coauther\_phy} \\
             \cline{2-13}&3& 10 &3 & 10 &3 &10 &3 & 10 &3 & 10&3 & 10 \\
             \midrule
             GCN  &0.615 &0.742 &0.505 &0.602 &0.583 & 0.652 &0.548 & 0.682 &0.800 & 0.893 & 0.851 &0.901\\
             InfoGNN-I  &0.683 &0.763 &0.582 &0.694 &0.599 & 0.739 &0.548 & 0.695 & 0.823 & 0.887 & 0.885 &0.928\\
             InfoGNN-IT  &0.696 &0.791 &0.589 &0.723 &0.619 & 0.768 
             &0.586 & 0.726 & 0.827 & 0.892 & 0.899 &0.941\\
             InfoGNN-ITS  &0.720 &0.792 &0.624 &0.728 &0.645& 0.766 
             &0.592 & 0.723 & 0.826 & 0.886 & 0.905 &0.941\\
             InfoGNN &0.737 &0.792 &0.654 &0.723 &0.669 & 0.765 
             &0.611 & 0.723 & 0.866 & 0.906 & 0.924 &0.942 \\
             \bottomrule
        \end{tabular}}
    \label{table:ablation}
\end{table*}
\vspace{-0.5cm}



\vspace{-2em}
\subsection{Ablation Study}
To further analyze how different components of the proposed method take effect, we conduct a series of ablation experiments. Due to space limit, we only report experimental results on the settings where 3 and 10 nodes are labeled per class. The ablations are designed as follows:
\begin{itemize}
    \item \textbf{InfoGNN-I}: only $\ell_I$ is applied based on GCN, which is used to evaluate the role of the contrastive loss;
    \item \textbf{InfoGNN-IT}: both $\ell_I$ and $\ell_T$ are applied, which is utilized to evaluate the impact of the GCE loss by comparing with InfoGNN-I. Note that only prediction score is applied here for $\ell_T$, i.e. $U_p = \{v \in U | f(\mathbf{x}_v)_j > k\}$;
    \item \textbf{InfoGNN-ITS}: on the basis of InfoGNN-IT, the informativeness score, i.e., Eq.(\ref{eq:up}), is also applied for $\ell_T$, which is to test the efficacy of the informativeness score by comparing with InfoGNN-IT. The impact of the $\ell_{KL}$ loss can be revealed by comparing with InfoGNN.
\end{itemize}
The ablation results are reported in Table~\ref{table:ablation}. From this table, we can see that each component of InfoGNN has its own contribution, but their contribution might differ at the two different settings. 
The constrastive loss seems to make similar contribution at the two settings, which achieves an average improvement of 3.6\% and 3.9\% over GCN on the six datasets. This proves that $\ell_I$ enables to produce node embeddings beneficial to node classification by maximizing the MI between nodes and their neighborhood. On top of $\ell_I$, when GCE is applied, model performance has been further boosted. Taking Dblp and Wikics as an example, it respectively boosts the accuracy by 1.6\% and 3.8 \% with 3 given labels per class, 2.9\% and 3.1\% with 10 given lables per class. This confirms the effectiveness of the GCE loss $\ell_T$. With regard to the CBR $\ell_{KL}$, it improves the performance by 2.48\% on average, with nearly 4\% on Coauther\_cs when 3 labels per class are given. Yet, we also find it contributes less on the setting of 10 given labels per class than that of 3, which is in line with our expectation. As in the situation of lower label rate, due to the limited label information, GNN is more prone to generate imbalanced predictions. Hence, the contribution of $\ell_{KL}$ is more remarkable. 
A similar consequence could also be seen on the contribution of candidate selection using representativeness scores. When there are few given labels, the selected nodes can bring in relatively more auxiliary information. In contrast, when more given labels are available, a larger amount of unlabeled nodes could be selected due to our slack pseudo-labeling constraints, which, to some extent, counteracts the effect of representative scoring. 


\subsection{Sensitivity Analysis}
\label{subsection: sensitivity}

We also conduct experiments to test the impact of hyperparameters ($\alpha, \beta, q$ and $k$) on the performance of InfoGNN. We take turns to test the effect of each hyperparameter while fixing the values of the rest. Due to space limit, we only report the results on Cora, Citeseer and Dblp when $3$ and $10$ labels per class are given for training.

\begin{figure}
    \centering
    \subfigure[$\alpha$ with 3 given labels per class]{
        \includegraphics[width=0.43\textwidth]{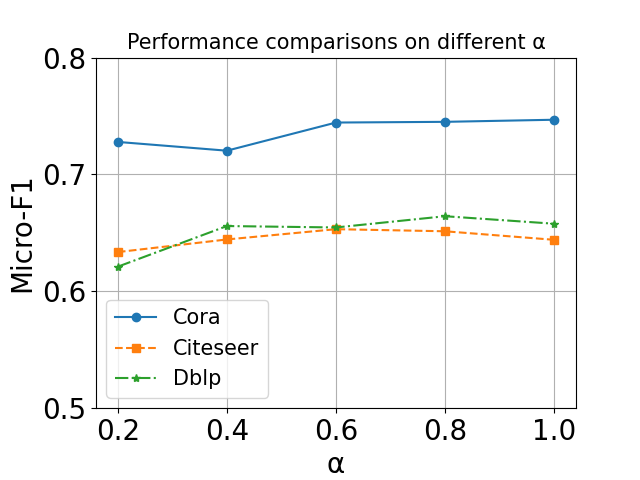}}
        \subfigure[$\alpha$ with 10 given labels per class]{
        \includegraphics[width=0.43\textwidth]{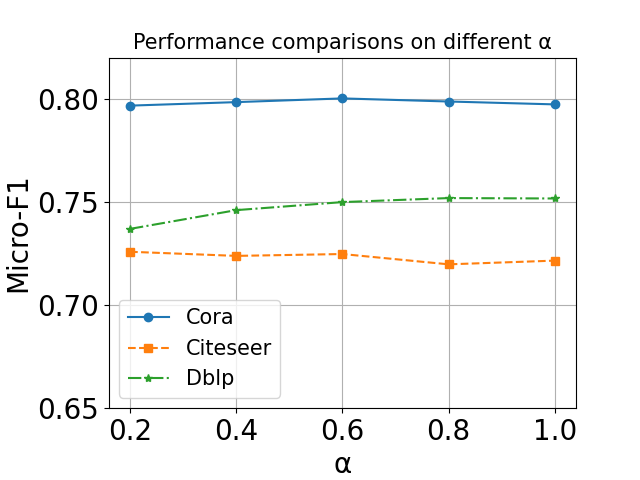}}
        \caption{Sensitivity analysis w.r.t. $\alpha$ on citation networks}
        \label{fig:alpha}
\end{figure}

Hyperparameter $\alpha$ controls the contribution of the contrastive loss $\ell_I$ to the total loss. Its impact on model performance is shown in Fig.~\ref{fig:alpha}. With 3 given labels per class provided, we find that a larger $\alpha$ could lead to better performance before $\alpha = 0.6$. After that, the performance retains at a good level with very slight changes. With 10 labels per class provided, except on Dblp, the changes of $\alpha$ do not largely impact model performance on Cora and Citeseer. This indicates that, when label information is very limited, our model requires stronger structural regularization to help generate discriminative node embeddings. On the other hand, when label information is relatively sufficient, network training is dominated by supervised loss from given labels. Thus, $\ell_I$ mainly takes effects when given labels are scarce. 

\begin{figure}
    \centering
    
        \subfigure[$\beta$ with 3 given labels per class]{
        \label{fig:beta3}
        \includegraphics[width=0.43\textwidth]{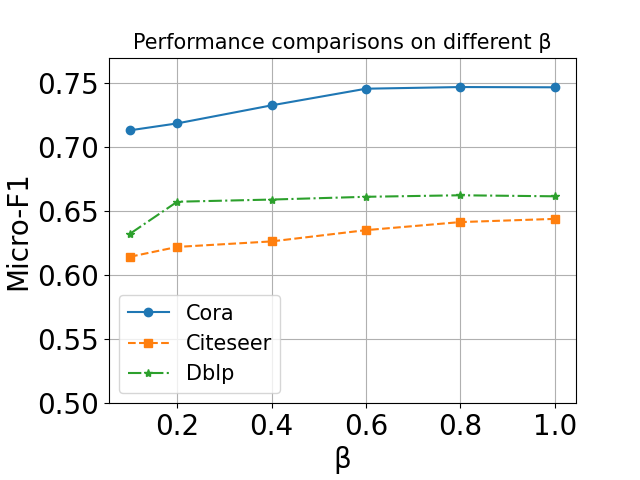}}
        \subfigure[$\beta$ with 10 given labels per class]{
        \includegraphics[width=0.43\textwidth]{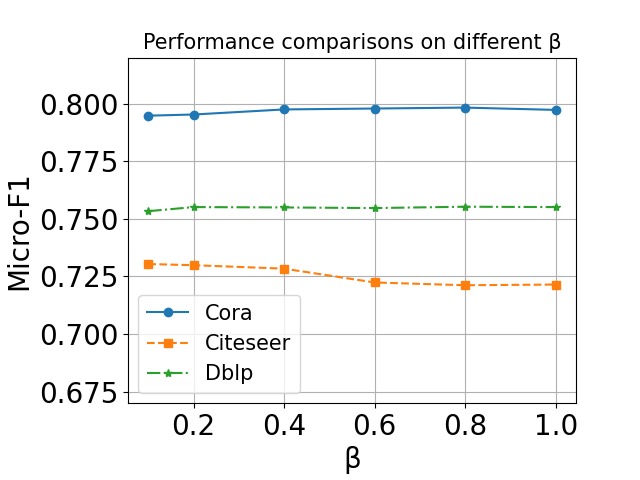}}
        
    \caption{Sensitivity analysis w.r.t. $\beta$ on citation networks}
    \label{fig:beta}
\end{figure}

Fig.~\ref{fig:beta} shows performance comparisons on different values of $\beta$. A similar trend with $\alpha$ can be observed on both settings. With only 3 labels per class provided, the class-imbalance problem is more likely to occur during pseudo labeling. Thus, our model favors a larger $\beta$ to regularize numbers of each pseudo label class to be relatively equivalent, as shown in Fig.~\ref{fig:beta3}. As $\beta$ increases from 0.1 to 1.0, our model boosts its classification accuracy by around 3\% on Citeseer and Cora. When 10 labels are given, as more label information can be exploited, the class-imbalance problem is less likely to arise. Hence, the change of $\beta$ does not result in much impact on  model performance.

\begin{figure}
    \centering
    \subfigure[$q$ with 1 given labels per class]{
    \label{fig:qw1}
        \includegraphics[width=0.43\textwidth]{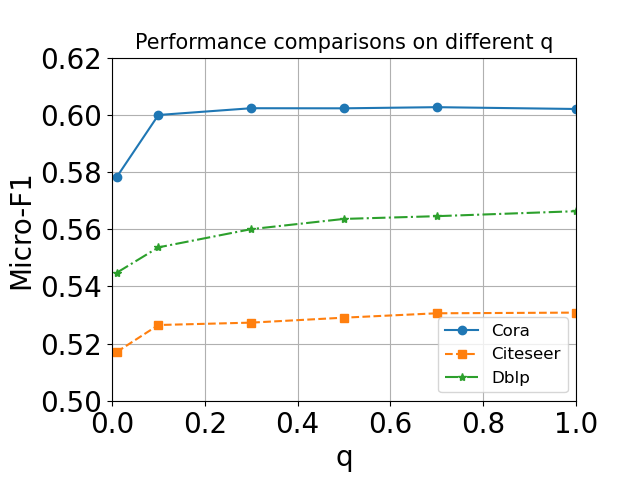}}
        \subfigure[$k$ with 1 given labels per class]{
        \label{fig:kw1}
        \includegraphics[width=0.43\textwidth]{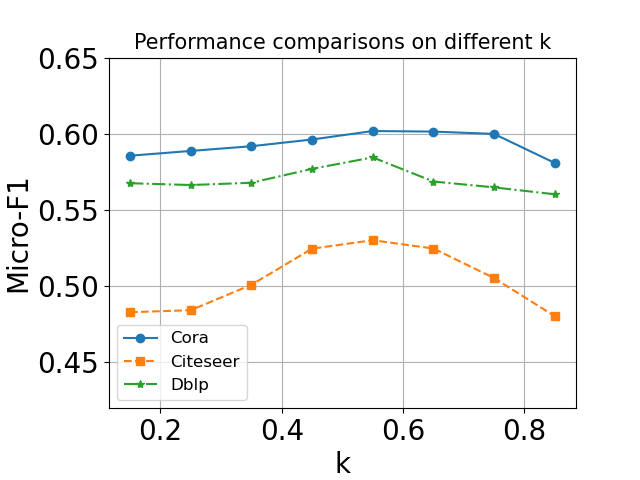}}
        
    \caption{Sensitivity analysis w.r.t. $q$ \& $k$ on citation networks}
\end{figure}

Hyperparameter $q$ is the generalization coefficient in $\ell_T$. 
Fig.~\ref{fig:qw1} illustrates model performance changes with an increase of $q$ when one label per class is given. We can see that, as $q$ rises, the performance of our method shows a gradual increase on the three datasets. This is because the severe lack of label information is more probable to incur noise in pseudo labels. A larger $q$ is then able to decay the gradient update on unreliable samples that have lower prediction probabilities. This reduces the sensitiveness of our model towards incorrect pseudo labels, leading to better performance. On the other hand, \textit{when descending $q$ near zero, the GCE loss is approaching close to SCE,} and at the same time, the model has a significant performance degradation. This further proves the superiority of GCE over SCE loss when only few labels are given for training. 

Hyperparameter $k$ is the threshold for $\ell_T$, which controls how many unlabeled nodes are selected for pseudo labeling. Fig.~\ref{fig:kw1} depicts the performance changes by varying $k$ with one given label per class. As we can see in this figure, a medium $k$ achieves better accuracy, while either too small or too large $k$ would undermine model performance. 


        

\vspace{-2em}
\section{Conclusion}
In this paper, we propose an informativeness augmented pseudo-labeling framework, called InfoGNN, to address semi-supervised node classification with few labels. We argue that all of the existing pseudo-labeling approaches on GNNs suffer from two major pitfalls: information redundancy and noisy pseudo labels. To address these issues, we design a representativeness measuring method to assess node informativeness based on MI estimation maximization. 
Taking both informativeness and prediction confidence into consideration, more informative unlabeled nodes are selected for pseudo labeling. We then propose a generalized cross entropy loss for pseudo labels to mitigate the negative effect of unreliable pseudo labels. Furthermore, we propose a class-balanced regularization in response to the potential class-imbalance problem caused by pseudo labeling. Extensive experimental results and ablation studies verify the effectiveness of our proposed framework, and demonstrate its superior performance to state-of-the-art baseline models, especially under very few-label settings.


\bibliographystyle{spmpsci}      
\bibliography{reference.bib}   


\end{document}